\documentclass[reprint,showpacs,superscriptaddress,amsmath,assume,aps,prx,floatfix,longbibliography]{revtex4-2} 
\usepackage{amsthm}
\usepackage{amsmath, mathtools}
\usepackage{amssymb,amsfonts,amsthm}
\usepackage[utf8]{inputenc}	
\usepackage[english]{babel}
\usepackage{amsmath}
\usepackage{graphicx}		
\usepackage{natbib}
\usepackage{textcomp}
\usepackage{gensymb}
\usepackage[usenames,dvipsnames,svgnames,table]{xcolor}
\usepackage[hidelinks,colorlinks=false,urlcolor=Cerulean,citecolor=black]{hyperref}
\usepackage{siunitx}
\usepackage{mathrsfs}
\usepackage{multirow}
\usepackage{bm}
\usepackage{xfrac}
\usepackage{comment}
\usepackage[normalem]{ulem}
\usepackage{bm}
\usepackage{steinmetz}

\renewcommand{\i}{\mathrm{i}}

\DeclareMathOperator{\R}{\mathbb{R}}
\DeclareMathOperator{\C}{\mathbb{C}}

\DeclareMathOperator\Arg{Arg}

\begin{document}

\title{Image segmentation with traveling waves in an exactly solvable recurrent neural network}

\author{Luisa H. B. Liboni}
\thanks{These authors contributed equally}
\author{Roberto C. Budzinski}
\thanks{These authors contributed equally}
\author{Alexandra N. Busch}
\thanks{These authors contributed equally}
\affiliation{Department of Mathematics, Western University, London, ON, Canada}
\affiliation{Western Institute for Neuroscience, Western University, London, ON, Canada}
\affiliation{Western Academy for Advanced Research, Western University, London, ON, Canada}
\author{Sindy Löwe}
\affiliation{AMLab, University of Amsterdam, Amsterdam, Netherlands}
\author{Thomas A. Keller}
\author{Max Welling}
\affiliation{UvA-Bosch Delta Lab, University of Amsterdam, Amsterdam, Netherlands}
\author{Lyle E. Muller}
\affiliation{Department of Mathematics, Western University, London, ON, Canada}
\affiliation{Western Institute for Neuroscience, Western University, London, ON, Canada}
\affiliation{Western Academy for Advanced Research, Western University, London, ON, Canada}

\begin{abstract}
We study image segmentation using spatiotemporal dynamics in a recurrent neural network where the state of each unit is given by a complex number. We show that this network generates sophisticated spatiotemporal dynamics that can effectively divide an image into groups according to a scene's structural characteristics. Using an exact solution of the recurrent network's dynamics, we present a precise description of the mechanism underlying object segmentation in this network, providing a clear mathematical interpretation of how the network performs this task. We then demonstrate a simple algorithm for object segmentation that generalizes across inputs ranging from simple geometric objects in grayscale images to natural images. Object segmentation across all images is accomplished with one recurrent neural network that has a single, fixed set of weights. This demonstrates the expressive potential of recurrent neural networks when constructed using a mathematical approach that brings together their structure, dynamics, and computation.
\end{abstract} 

\maketitle

Image segmentation is a fundamental task in computer vision. Whether finding regions of interest in medical images or highlighting specific objects, the ability to effectively divide an image into groups based on the structure in a scene can greatly facilitate image processing. Many techniques have been developed for image segmentation, from classical watershed~\cite{beucher1979use} or active contour~\cite{kass1988snakes} algorithms to modern slot-based~\cite{hinton2018matrix} and deep-learning approaches~\cite{minaee2021image}. Automated object segmentation represents a specific goal for image segmentation algorithms, in which pixels within the same object are grouped together. Finding objects in a data-driven manner allows dividing an input image into parts, opening up opportunities for further processing and semantic understanding.

Recent work in unsupervised image segmentation has utilized a special kind of autoencoder, where each node in the network has a state characterized by a complex number~\cite{lowe2022complex}. Because the state of each node in this complex-valued network has both an amplitude and a phase, the intensity of each pixel can be encoded in the amplitude, and objects can be encoded in groups of nodes with similar phases. In terms of a physical system, groups of nodes with similar phases correspond to oscillators that synchronize when they are part of the same object. This physical analogy has been utilized in segmenting images in networks of spring-mass harmonic oscillators \cite{belongie1998finding}, chaotic maps~\cite{zhao2001network,zhao2003network}, and Kuramoto oscillators \cite{ricci2021kuranet}, where nodes that are part of the same object will synchronize to a phase that is unique from the phase of different objects.

In these previous works, segmentation occurs when all oscillators within an object synchronize on approximately the same phase. Beyond complete synchrony, however, networks of nonlinear oscillators can also display sophisticated spatiotemporal patterns. These patterns include ``chimera'' states~\cite{abrams2004chimera,panaggio2015chimera}, where only a pocket of nodes is synchronized and the others are desynchronized, and traveling waves~\cite{laing2016travelling,zanette2000propagating,jeong2002time}. In neuroscience, traveling waves have recently been found in recordings from the visual cortex during active sensory processing \cite{muller2018cortical}. Specifically, in studying the visual cortex during active visual processing, we have found that a small visual stimulus evokes a wave of activation traveling outward from the point of input~\cite{muller2014stimulus} and that natural image stimuli also evoke waves of activity traveling over an entire cortical region~\cite{davis2020spontaneous}. 

These waves traveling over individual cortical regions prompt an interesting computational question. Input from the eyes is organized into a retinotopic map \cite{swindale2008visual}, in which neurons at a single point in a visual region receive feedforward input from only a small portion of visual space, and each cortical region in the visual system contains an entire map of the visual field. This suggests that visual processing is relatively local and that a static image input would result in a static activity pattern, corresponding to how well local patches of the image drive the feature selectivity in each cortical region \cite{riesenhuber1999hierarchical}. The feedforward input, however, represents only approximately 5\% of the inputs to a single cell in visual cortex \cite{markov2011weight}. While these feedforward synapses are strong, the dense, within-region recurrent connectivity makes up about 80\% of connections a cell receives \cite{markov2011weight}. We have recently found that this recurrent connectivity generates traveling waves in single regions of visual cortex \cite{muller2014stimulus,davis2021spontaneous}, potentially overlaying the input to individual cortical regions with additional, internally generated dynamics. What could be the computational advantage of these internally generated traveling waves? Could they interfere with local cortical processing, or could they perhaps provide some computational benefit?

In this work, we demonstrate that networks of oscillators can perform object segmentation with traveling waves. We focus on oscillator networks with recurrent architecture, where nodes within the same layer can be connected. This recurrent architecture, which is consistent with the dense connectivity found in visual cortex, is distinct from the standard feedforward networks (where nodes are arranged into layers with no intra-layer connections) often employed for these tasks \cite{minaee2021image}. In comparison, recurrent neural networks (RNNs) are considered relatively less often for object segmentation tasks, in part because they are more difficult to train than feedforward networks \cite{bengio1993problem, pascanu2013difficulty}.

We have recently developed a line of research that uses spectral graph theory to gain insight into the dynamics of recurrent oscillator networks \cite{budzinski2022geometry,budzinski2023analytical}. Using the insights gained from this line of work, we introduce a recurrent network model that can perform object segmentation with internally generated spatiotemporal dynamics. Nodes in this network have a state defined by a complex number, with an amplitude and a phase. We find that this complex-valued recurrent neural network (cv-RNN) can segment objects in input images with very simple linear interactions between nodes. Combining insights derived from spectral graph theory and our mathematical analysis of nonlinear oscillator networks, we design a linear cv-RNN that exhibits long transients in each node's amplitude, while also exhibiting meaningful evolution of the phases. While amplitudes can diverge in linear systems, the fact that the dynamics of phase remain bounded represents a potential advantage of complex-valued linear systems for computation. This approach makes it possible to leverage the simplicity of a linear network while also keeping the dynamics bounded in a range that is useful for processing visual inputs. In addition, using complex-valued networks in this context will simplify the equations used later in the mathematical analysis. We find that this network can segment simple geometric objects in binary images, mixtures of geometric objects and greyscale MNIST digits \cite{lecun1998mnist}, and naturalistic images while also being exactly solvable. We then present a complete mathematical analysis of how the cv-RNN performs this segmentation, using the fact that we can solve the equations for the network dynamics exactly.

These results open a novel avenue in image segmentation by providing fundamental insight into how oscillator networks \cite{belongie1998finding,ricci2021kuranet} and complex-valued autoencoders \cite{lowe2022complex} can be trained to perform object segmentation. These results also open possibilities for new algorithms and hardware-based implementations because linear complex-valued networks are easy to implement directly in electronics. These results are consistent with recent observations that linear RNNs may have key advantages over Transformers in some long sequence prediction tasks \cite{orvieto2023universality,orvieto2023resurrecting}, while our results also extend the applicability of these RNNs to image processing tasks. In this work, however, we have also simplified the network architecture enough to make the recurrent dynamics exactly solvable, opening up new paths for the mathematical analysis of functioning neural networks. Finally, these results also provide insight into visual processing in biological brains, by demonstrating a first computational example of why populations of neurons in a region of visual cortex might respond to a {\it static} stimulus with a {\it dynamic} activity pattern, as we have recently observed in experimental recordings.

\subsection*{Network architecture}

\begin{figure}[b]
    \centering
    \includegraphics[width=\columnwidth]{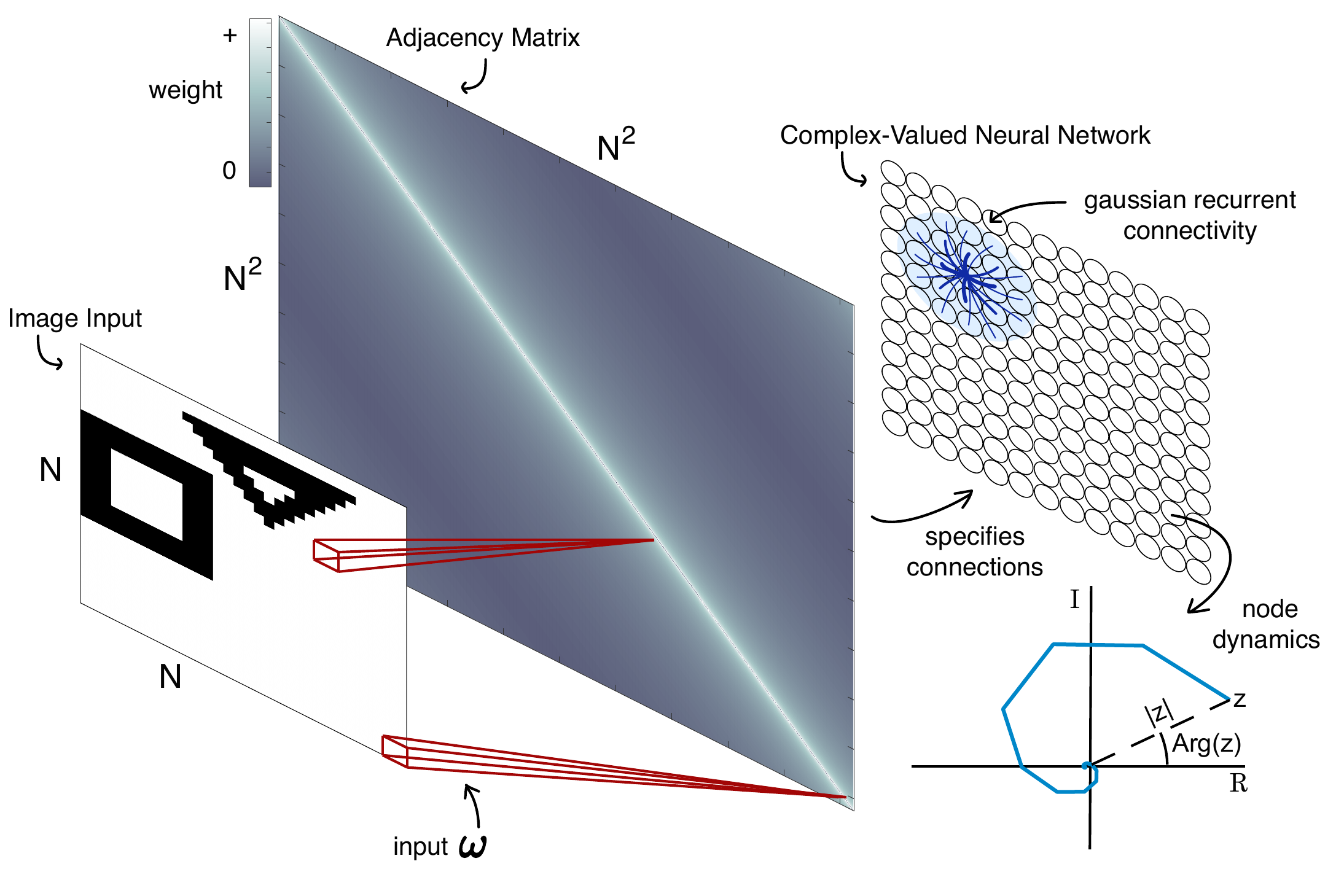}
    \caption{
    \textbf{Schematic representation of the cv-RNN.} The network is composed $N^2$ nodes (top right). The activity of each node is described by a phase $\mathrm{Arg}(z)$ and an amplitude $|z|$ in the complex plane (bottom right). Nodes are placed in a 2-dimensional sheet, where recurrent connection weights (blue lines, right) decrease as a Gaussian with distance between nodes Eq.~(\ref{eq.gaussian}), defining a weighted adjacency matrix with $N^{2} \times N^{2}$ entries (center). An image represented in a 2-dimensional sheet with $N \times N$ pixels is then flattened into a vector $\bm{\omega}$ with $N^{2}$ entries. The image input modulates the intrinsic frequency of each node in the recurrent network. To do this, the vector $\bm{\omega}$ is added to the diagonal entries of the adjacency matrix, to form a composed matrix $\bm{B}$ that represents the recurrent connections and the input.}
    \label{fig:fig1}
\end{figure}

The cv-RNN is arranged on a two-dimensional square lattice with a side length of $N$ nodes. Each node in the network receives input from one pixel of an image (Fig.\,\ref{fig:fig1}). Nodes in the oscillator network are densely connected with their local neighbors (blue lines, ``Gaussian recurrent connectivity'', Fig.\,\ref{fig:fig1}), approximately following the connectivity that occurs in single regions of visual cortex~\cite{markov2011weight}. We consider a specific dynamical equation for the evolution of this system of $N^2$ nodes:
\begin{multline}
\label{eq.1}
    \dot {\psi_{i}}(t) = \omega_i + \epsilon \sum_{j = 1}^{N^2} a_{ij}  \left[ \right. \sin \left( \psi_j(t) - \psi_i(t) \right) \\ 
    - \i \, \cos \left(\psi_j(t) - \psi_i (t)\right) \left. \right],
\end{multline}
where $\psi_t(t) \in \C$ is the state of node $i$ at time $t$, $\omega_i \in \R$ is the node's intrinsic oscillation frequency, the matrix element $a_{ij} \in \R$ is the connection between nodes $i$ and $j$, and $\epsilon \in \R$ scales the strength of all connections. We note that throughout this paper, we consider $\i$ the imaginary unit, such that $\sqrt{-1} = \i$, in contrast to $i$, which represents an index.

\begin{figure*}[t!]
    \centering
    \includegraphics[width=0.8\textwidth]{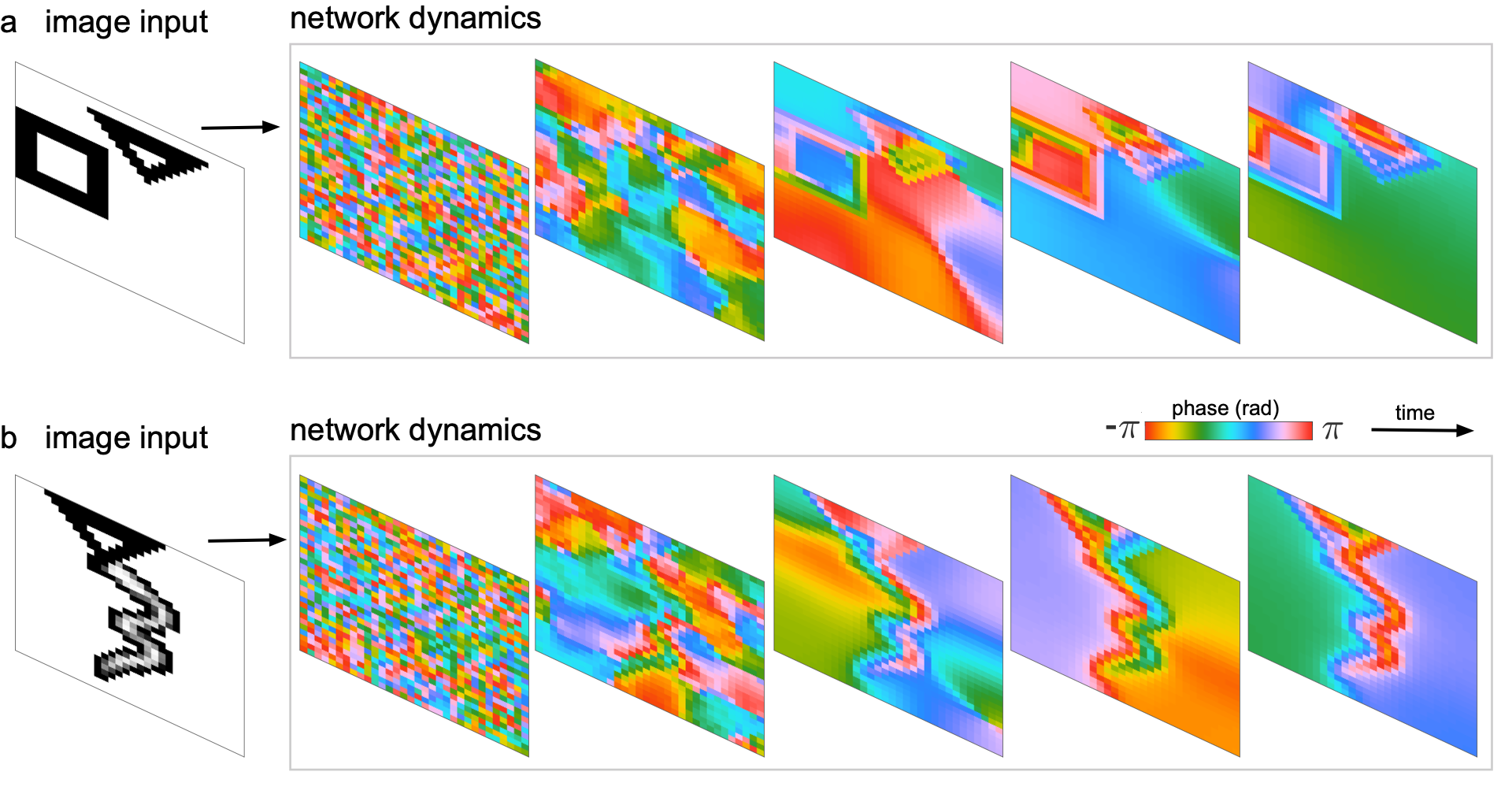}
    \caption{\textbf{Spatiotemporal dynamics produced by the cv-RNN}. \textbf{(a)} An image drawn from the 2Shapes dataset (see Methods, Visual inputs, and data set) is input to the oscillator network by modulating the nodes' intrinsic frequencies $\bm{\omega}$. The samples of the phase dynamics in the recurrent layer during transient time show that the nodes are imprinting the visual space by generating three different spatiotemporal patterns: one for the nodes corresponding to the background in the input space, one for the nodes corresponding to the square in the input image, and lastly for the nodes corresponding to the triangle in the input space.
    \textbf{(b)} Image drawn from the MNIST\&Shapes dataset is input into the dynamical system. Three different spatiotemporal patterns arise: one for the nodes corresponding to the background in visual input space, one for the nodes corresponding to the triangle in the input space, and lastly for the nodes corresponding to the handwritten three-digit.
    }
    \label{fig:fig2}
\end{figure*}
%

\begin{figure*}[t]
    \centering
    \includegraphics[width=\textwidth]{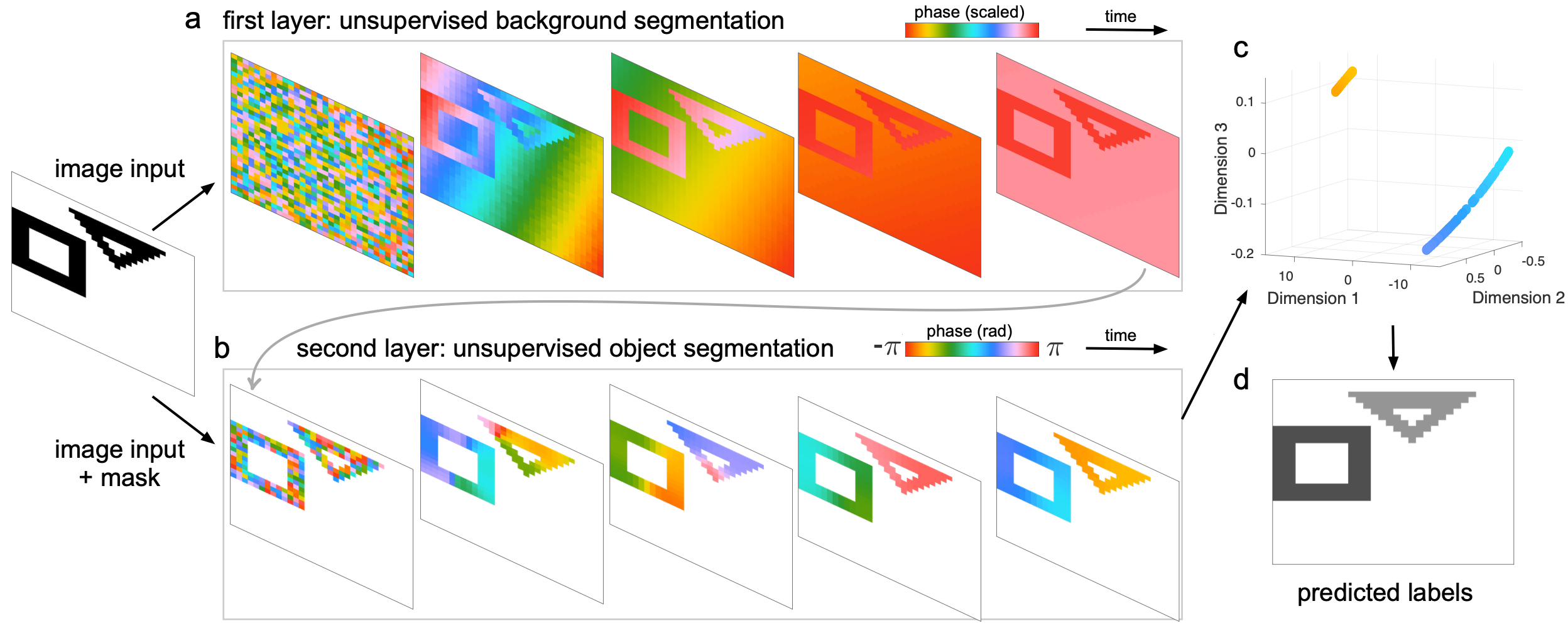}    
    \caption{\textbf{Object segmentation algorithm}. \textbf{(a)} A first cv-RNN layer with broad spatial connectivity segments the image background. In this plot, samples of the phase dynamics (reshaped to a $N \times N$ grid) at each point in time show a unique phase for the nodes corresponding to the background in the visual input space. Pixels corresponding to foreground objects synchronize on a single phase distinct from the background. \textbf{(b)} After timestep $k = 60$, nodes corresponding to background pixels are disconnected from the rest of the recurrent network in the second cv-RNN layer. Then, the second layer dynamics begins, where connections between nodes in the second layer create sophisticated spatiotemporal dynamics unique to each object. \textbf{(c)} The similarity projection in the low dimensional space for the phase dynamics generated in the second layer shows that the spatiotemporal patterns propagate through the nodes corresponding to the objects of the visual input. The phase patterns are separated into two different groups by the K-means algorithm.  \textbf{(d)} Labels assigned to objects in the input by the K-means algorithm.}
    \label{fig:fig3}
\end{figure*}

We have recently demonstrated that this specific nonlinear system displays the hallmark behaviors of synchronization that have been found in oscillator networks \cite{budzinski2022geometry,budzinski2023analytical}. Further, by defining the change of variable,
\begin{align}
\begin{split}
    x_i(t) &= e^{\i \psi_i(t)} = e^{-\operatorname{Im} (\psi_i(t))} e^{\i \operatorname{Re} (\psi_i(t))}\\ & =   | x_i(t)| \,  \Arg{ [x_i(t)]},
    \end{split}
\end{align}
we find that Eq.\,\eqref{eq.1} admits an exact solution \cite{muller2021algebraic,budzinski2022geometry}. Now, taking the system in discrete time, we can express the solution as
\begin{align}
    \label{eq.xk+1 = A x}
    \bm{x}(k+1) &= \underbrace{( \operatorname{diag} ( \i \bm{\omega} ) + \epsilon \bm{A}  ) }_{\bm{B}}\bm{x}(k),
\end{align}
in matrix form (see Supplementary Material, Sec.~I), where $\bm{A} \in \R^{N^2 \times N^2}$ contains the connections in the network. Here, we input the image to the recurrent network by modulating the intrinsic frequency of each node. Specifically, each pixel of the image drives the intrinsic frequency $\omega_i$ of each node, with higher pixel intensities resulting in faster oscillation frequencies. The matrix $\bm{B} \in \C^{N^2 \times N^2}$ describes the complete cv-RNN, where image inputs interact with recurrent connections to produce spatiotemporal patterns that allow segmentation.

Connections in the recurrent layer have strength that decreases with their Euclidean distance $d_{ij}$ between two nodes on the square lattice:
\begin{equation}
\label{eq.gaussian}
a_{ij} = \alpha \, \exp\left (\frac{-d_{ij}^2}{2\sigma^2}\right),
\end{equation}
where $\alpha \in \R$ sets the peak strength of connections, and $\sigma \in \R$ controls how fast connection strength falls off with distance. The architecture of these connections sets the scale for local recurrent interactions in the cv-RNN, allowing nodes whose input pixels are nearby the opportunity to interact and create shared spatiotemporal patterns. Throughout this work, the cv-RNN starts with random initial conditions, with node amplitudes $|x_i(0)|$ distributed uniformly in the interval $[0, 1]$ and phases $\Arg{[\bm{x}(0)]}$ uniform in $[-\pi, \pi]$.

\subsection*{The cv-RNN creates spatiotemporal patterns unique to each object in an input image}

We first consider the cv-RNN with inputs drawn from a dataset used in recent work on complex-valued autoencoders \cite{lowe2022complex}, which have simple, binary-encoded geometric shapes. We study the cv-RNN dynamics on these simple geometric objects, and on combinations of geometric objects and MNIST digits, before moving to more complex inputs and naturalistic images. With an input containing a triangle and a square, the cv-RNN begins with random initial conditions, where the phases of the nodes are desynchronized (Fig.\,\ref{fig:fig2}a, ``network dynamics'', first panel). Interactions between nodes are captured by elements of the system matrix $b_{ij}$, where the absolute value of the connection $|b_{ij}|$ changes the strength of interaction between two nodes, and the phase of the connection $\Arg{[b_{ij}]}$ changes their relative angle. These features of the connections are sufficient to drive traveling waves unique to the triangle, square, and image background (Fig.\,\ref{fig:fig2}a, ``network dynamics''; see also Supplementary Movie 1). Importantly, the wave traveling over the background has a substantially lower spatial frequency than the waves traveling over each object in the image, separating the objects and background into two very different sets of spatiotemporal patterns. With the same set of recurrent weights $\bm{A}$, and a new input image -- this time containing a triangle and an MNIST numeral $3$ -- the cv-RNN again produces unique waves traveling over the triangle, the ``3'', and the background (Fig.\,\ref{fig:fig2}b). These results demonstrate that the cv-RNN can generate spatiotemporal patterns from its internally generated recurrent dynamics. We next tested whether these spatiotemporal dynamics, in combination with an unsupervised method for separating the individual phase patterns we have recently developed \cite{M1}, could perform image segmentation.

\begin{figure*}[t]
    \centering
    \includegraphics[width=\textwidth]{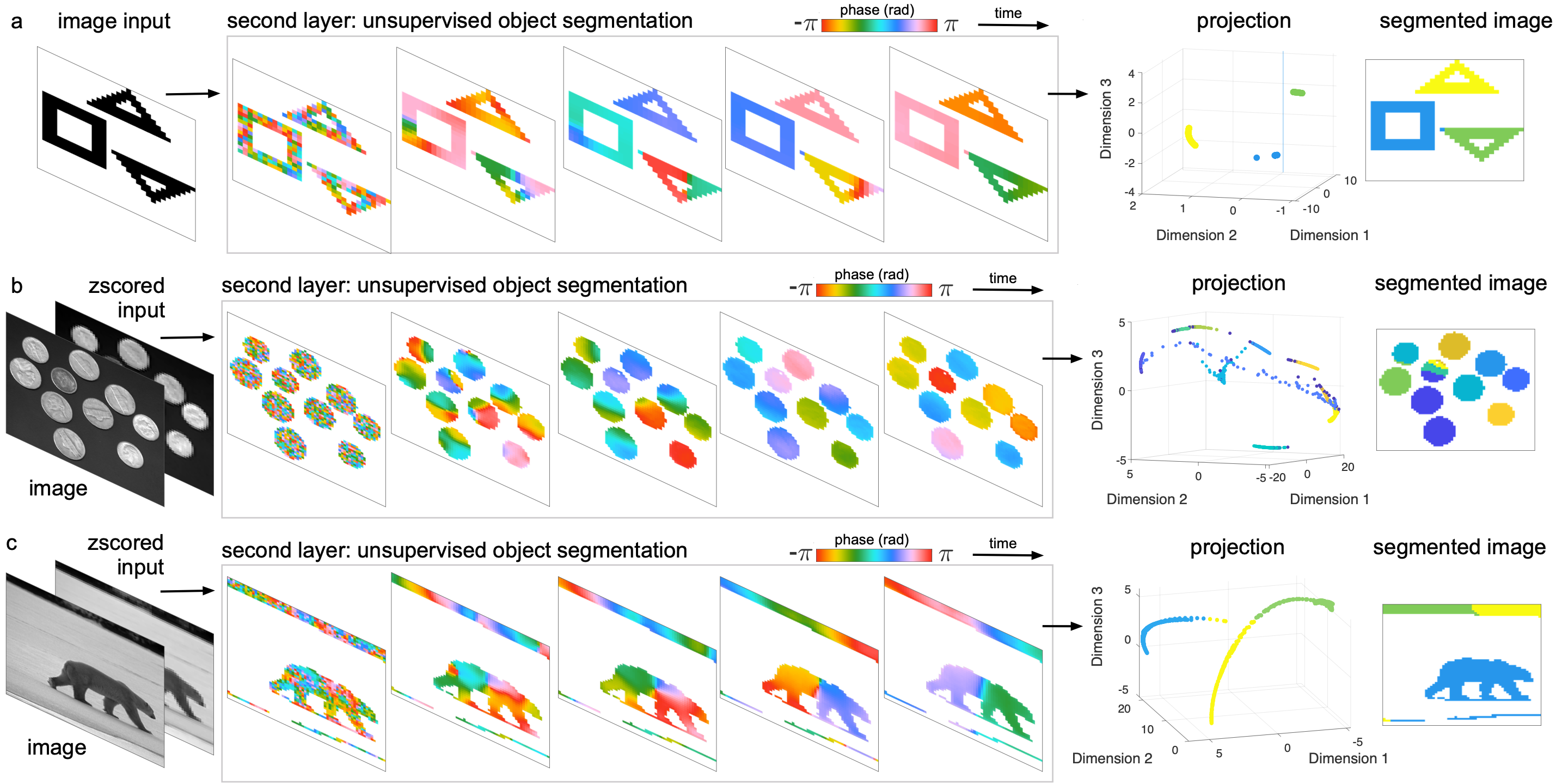}
    \caption{\textbf{A single set of recurrent connection weights segments objects from simple images to naturalistic visual scenes.} Image inputs are shown in the left column, and some samples of the phase dynamics are resized to a $N \times N$ grid and depicted in the middle column. Panel (a) contains an input with simple geometric shapes, panel (b) shows a naturalistic image input of coins on a dark background and panel (c) also shows a naturalistic image of a bear. Projection onto the eigenvectors of the similarity matrix separates the phase patterns in a three-dimensional space (second column from right, labeled ``projection''). Labels assigned to objects in the input by the K-means algorithm are plotted in the right column.}
    \label{fig:fig4}
\end{figure*}

\subsection*{Object segmentation algorithm}

Having observed that recurrent interactions can produce traveling wave patterns unique to each object, we next developed an algorithm to segment objects using these dynamics. This algorithm uses a two-layer implementation of the cv-RNN, where the first layer separates image objects from the background. Briefly, after a fixed number of time steps, the dynamics in the first layer separate the objects in an image from the background. This separation then determines the recurrent connectivity between nodes in the second layer, whose dynamics are run in order to segment the individual objects. In this way, the algorithm comprises a two-step approach to object segmentation, with each step solved through linear dynamics in a cv-RNN.

Connection patterns specific to each layer facilitate this process. In the first layer, the recurrent connections have a higher peak strength $\alpha$ and a broader spatial scale $\sigma$. The broad spatial scale of the recurrent connections, together with the different intrinsic frequencies driven by the input, creates a difference in the dynamics for nodes whose inputs have objects as input and for those nodes whose input is the background. With this architecture in the first layer, the phase dynamics converge to two different sets of synchronous nodes, with one set capturing the background and the other capturing the objects (Fig.\,\ref{fig:fig3}a). We then segment the background through a simple thresholding procedure. In the second layer, nodes assigned to the background are disconnected from the rest of the recurrent layer, and the remaining recurrent connections have lower $\alpha$ and a smaller spatial scale $\sigma$. With this architecture in the second layer, the phase dynamics of the cv-RNN then display traveling waves that clearly separate the individual objects in the image (Fig.\,\ref{fig:fig3}b). By conducting a comprehensive numerical study over network hyperparameters for the training images in the 2Shapes dataset \cite{lowe2022complex}, we were able to identify a single set of recurrent weights for layers 1 and 2 that generates clearly unique traveling wave patterns over image objects, across cases where the objects are in different positions in the image and at different relative distances.

The only remaining step is to segment the phase patterns in the second layer into specific object labels. To do this, we use a method we have recently developed to find repeated spatio-temporal patterns in multisite neural recordings \cite{M1}. Briefly, if $\bm{\phi}_i$ represents the phase dynamics of node $i$ for a set of time points $\mathcal{T}$, then we can compute similarity $s_{jk}$ between nodes $j$ and $k$ through the complex inner product:
\begin{equation}
    s_{jk} = \frac{1}{T} \langle \bm{\phi}_j, \bm{\phi}_k \rangle ,
\end{equation} 
where $T = | \mathcal{T}|$. By computing the similarity between the dynamics of each pair of nodes, we can then construct a similarity matrix $\bm{S} \in \C^{N^2 \times N^2}$. Because $\bm{S}$ is complex-valued and Hermitian, its eigenvalues $\lambda_1, \lambda_2, ... , \, \lambda_N$ are real-valued, and its eigenvectors ${ \bm{z}_1, \bm{z}_2, ... , \, \bm{z}_N}$ have complex elements. Note that we will order the eigenvalues by increasing absolute value, so that $ | \lambda_1 | \geq | \lambda_2 | \geq ... \geq| \lambda_N |$. We then project the real part of $\bm{S}$ onto the real part of its leading eigenvectors:
\begin{equation}
\bm{P} = \hat{\bm{S}} \, \hat{\bm{Z}},
\end{equation}
where $\hat{\bm{S}}$ is the element-wise real part of $\bm{S}$, and $\hat{\bm{Z}}$ is a matrix whose columns are the real part of eigenvectors $\bm{z}_1$, $\bm{z}_2$, and $\bm{z}_3$. This projection defines a three-dimensional space that describes the similarity between phase dynamics. Each node in the cv-RNN becomes one point in this space, with node position determined by its relative similarity to the dynamics of the other nodes over time. Clustering the individual waves traveling over each object is straightforward in this three-dimensional similarity space (Fig.\,\ref{fig:fig3}c). A simple K-means clustering algorithm using the points in this similarity space then produces correct object labels (Fig.\,\ref{fig:fig3}d). We use K-means on this similarity space throughout this work but note that more sophisticated clustering algorithms can be applied in future work. 

With this approach, the two-layer cv-RNN robustly segments objects in the set of inputs with two geometric objects in the test dataset \cite{lowe2022complex}. On average, 93\% of pixels in 1000 images of two non-overlapping geometric shapes were correctly clustered, and 86\% in 1000 images of three non-overlapping geometric shapes (see also Supplementary, Sec. III), comparable to the range reported in previous work \cite{reichert2014, locatello2020, lowe2022complex}. These results demonstrate that the cv-RNN developed here enables generalization to inputs where objects are not in the same position of the image, but can have rotation or translation.

\begin{figure*}[t]
    \centering
    \includegraphics[width=\textwidth]{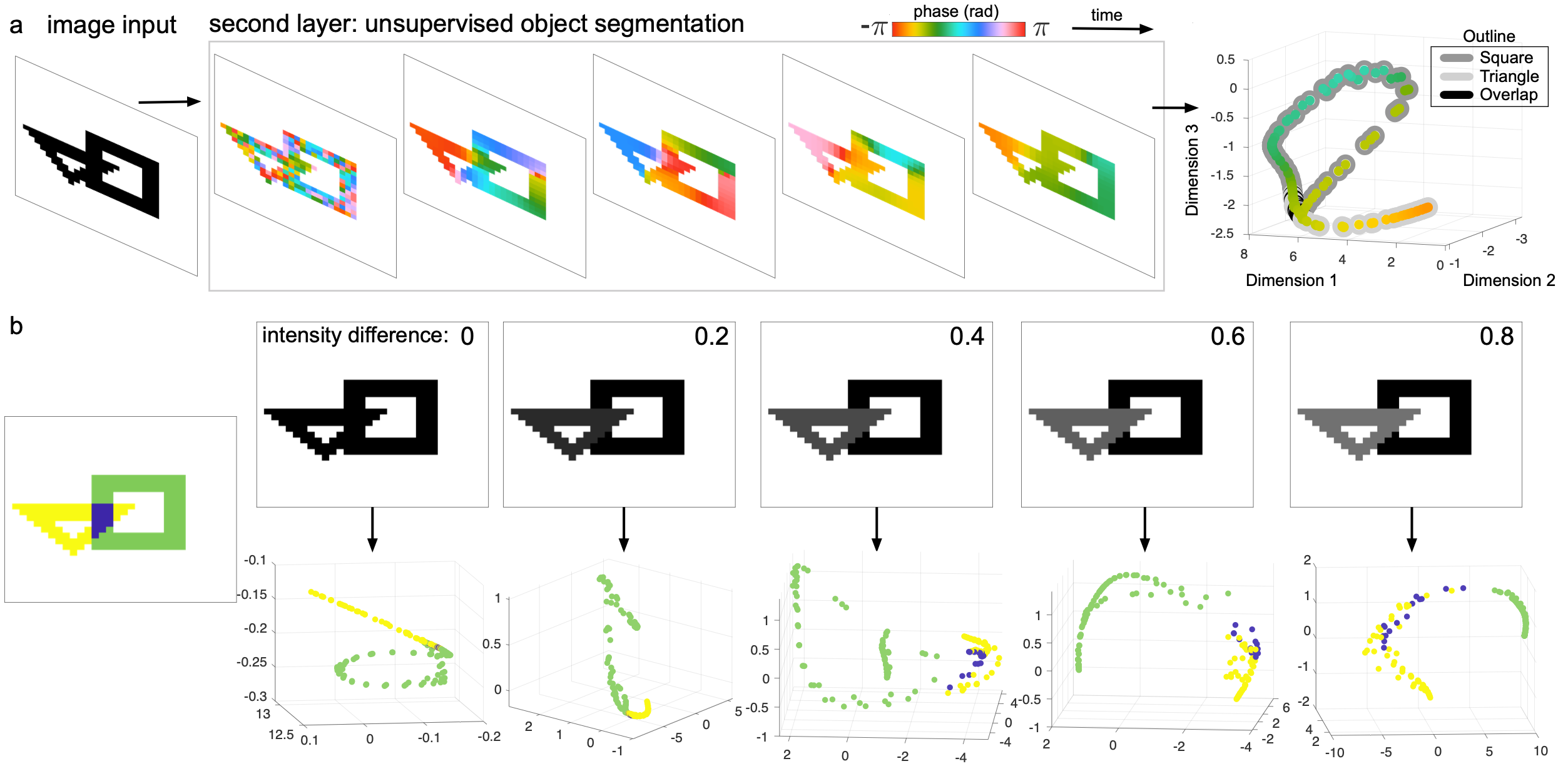}
    \caption{\textbf{Segmentation of overlapping objects.} \textbf{(a)} 
    When objects in the input image overlap, the object segmentation algorithm separates the non-overlapping sections into different objects. (3D plot at right) Points in the projection are colored by the final value in the phase dynamics. Outlines for each point denote the object to which each point belongs in the ground-truth input. The points are arranged into two closed loops that meet where the pixels in the image overlap. \textbf{(b)} Small differences in the pixel intensities for each object separate the similarity projection for each object. (left) Ground-truth labels for this case of partial overlap, with the triangle in yellow, the square in green, and pixels belonging to the overlap zone in purple. (top row) Differences in pixel intensities for each foreground object range from 0 to 0.8 (top right corner of input images), and nodes in the overlap zone (purple nodes) receive the same input intensity as the triangle (yellow nodes). (bottom row) Plotted is the similarity projection for each input case, with nodes in the projection colored according to which zone they belong in the input image. When the pixel intensities differ for the two objects, the pixels in the overlap zone are assigned the intensity of the triangle. As the difference in pixel intensity between the triangle and the square increases, the separation between clusters in the similarity projection grows. As in Fig.~\ref{fig:fig4}, all image segmentation is performed with the same set of recurrent weights and hyperparameters.}
    \label{fig:fig5}
\end{figure*}

\subsection*{A single set of recurrent weights segments images across datasets}

Having developed a two-layer cv-RNN that can perform object segmentation in simple images, we next tested the cv-RNN for more complex inputs and natural images. Using the same set of recurrent weights identified above, we find that the cv-RNN can also perform object segmentation on these more sophisticated examples. The cv-RNN can successfully segment inputs with three or more distinct geometric objects in the image (Fig.\,\ref{fig:fig4}a). Further, and again using the same set of recurrent weights derived from the simpler image sets, the cv-RNN can also segment natural images through these two-layer recurrent dynamics (five natural images total, see Fig.\,\ref{fig:fig4}b,c and Supplementary Fig. S2). One image with ten coins (Fig.\,\ref{fig:fig4}b) and one image with a bear (Fig.\,\ref{fig:fig4}c) are z-scored and input directly into the cv-RNN. Even with these more sophisticated natural inputs, unique traveling waves occur for each object in the second layer, allowing segmentation of all coins in the input image. 

These results demonstrate that the cv-RNN can generalize to inputs with different numbers of objects, and even to novel visual inputs, without changing weights or hyperparameters, demonstrating the breadth of inputs that can be handled by the recurrent dynamics in our approach. We note that the fixed set of weights will depend on the spatial scale of the objects in the input; however, we can use our approach for segmentation with multiple object sizes through a hierarchy of layers with different distance connectivity scales. 

In these previous cases, the images considered contained sets of objects that were non-overlapping. Phase dynamics within the second recurrent layer create unique traveling wave patterns that eventually converge to unique synchronized phases for each object. The interaction of the random initial conditions with the recurrent interactions in the second layer allows the network to generate different phase values for the nodes corresponding to different objects in the image input. However, an important question is whether this approach can also work when objects overlap to some extent. An example of this case is an image of a triangle and a square from the binary shapes dataset \cite{lowe2022complex} where the two objects overlap. With this input, the network produces two distinct spatiotemporal patterns for each object (Fig.\,\ref{fig:fig5}a; see also Supplementary Movie 5). Nodes receiving input from the triangle exhibit a wave traveling in a counter-clockwise direction across the object, while nodes for the square exhibit a wave traveling in the clockwise direction. The pixels where the two objects do not overlap can then be segmented using the unsupervised phase similarity method (Fig.\,\ref{fig:fig5}a, right). The similarity projection clearly reflects the structure of the dynamics, with the pixels in each object forming closed loops that meet where the pixels in the image overlap. In this overlap zone, the pixels display an interesting behavior, as the two waves meet in the overlap zone, and these nodes then exhibit phases that are consistent with either spatiotemporal pattern (red nodes, Fig.\,\ref{fig:fig5}a). This ambiguity, however, only holds for the case where inputs are binary. In the case where pixel intensities differ slightly for each object, as expected for natural images in general, the phase similarity representation shows increasing separation for each object as the difference in intensity grows (Fig.\,\ref{fig:fig5}b). These results hold over a range of inputs with partially overlapping objects (Supplementary Material, Sec.\,IV). 

These results demonstrate that the cv-RNN produces unique spatiotemporal patterns that enable segmentation with our unsupervised phase similarity technique, even in the non-trivial case where objects in the input overlap. The cv-RNN tolerates substantial overlap before the phase patterns become indistinguishable (Supplementary Material, Sec. IV, Fig.\,S1). Finally, it is important to note that, while we do not consider segmentation for the case of complete object overlap here, and instead focus on the cases where objects can be separated based on information present in the individual input image, adding additional learning mechanisms to the cv-RNN in future work could allow the network to identify specific objects for segmentation, in addition to performing more sophisticated image processing tasks.

\subsection*{The exact solution for the cv-RNN allows mathematical analysis of the segmentation computation}

In addition to segmenting objects in input images with a single set of recurrent weights, the cv-RNN introduced in this work is unique because it admits an exact mathematical solution. This means that we can precisely analyze how the recurrent layers perform their computations. Because a recurrent layer is the linear dynamical system in Eq.~\eqref{eq.xk+1 = A x}, it can be described by means of its eigenvalues and eigenvectors as follows: 
\begin{equation}
\bm{x}(k) = \bm{B}^k \,  \bm{x}(0) = \sum_{i= 1}^{N^2} \underbrace{\lambda_{i}^k \left( \,  \bm{r}_{i}^T \bm{x}(0) \,  \right)}_{\mu_{i}(k)} \bm{v}_{i}
\label{eq.dyn},
\end{equation}
where $\lambda_{i}$ are the eigenvalues associated with eigenvectors $\bm{v}_{i}$ of $\bm{B}$, $\bm{r}_{i}^{T}$ are the rows of $[\bm{v}_{1} \cdots \bm{v}_{{N^2}}]^{-1}$, and coefficients $\mu_{i}(k)$ weight the contribution of each eigenvector.

The linear combination of eigenvectors generates the spatiotemporal patterns that arise during the transient dynamics of the cv-RNN. The contribution of each eigenvector is weighed by the coefficient $\mu_{i}(k)$, which is dependent on the initial condition and the eigenvalues. Each contribution will scale and rotate the eigenvectors of the nodes at each timestep $k$, giving rise to the spatiotemporal dynamics on the arguments of $\bm{x}(k)$.
\begin{figure}[t!]
    \centering
    \includegraphics[width=\columnwidth]{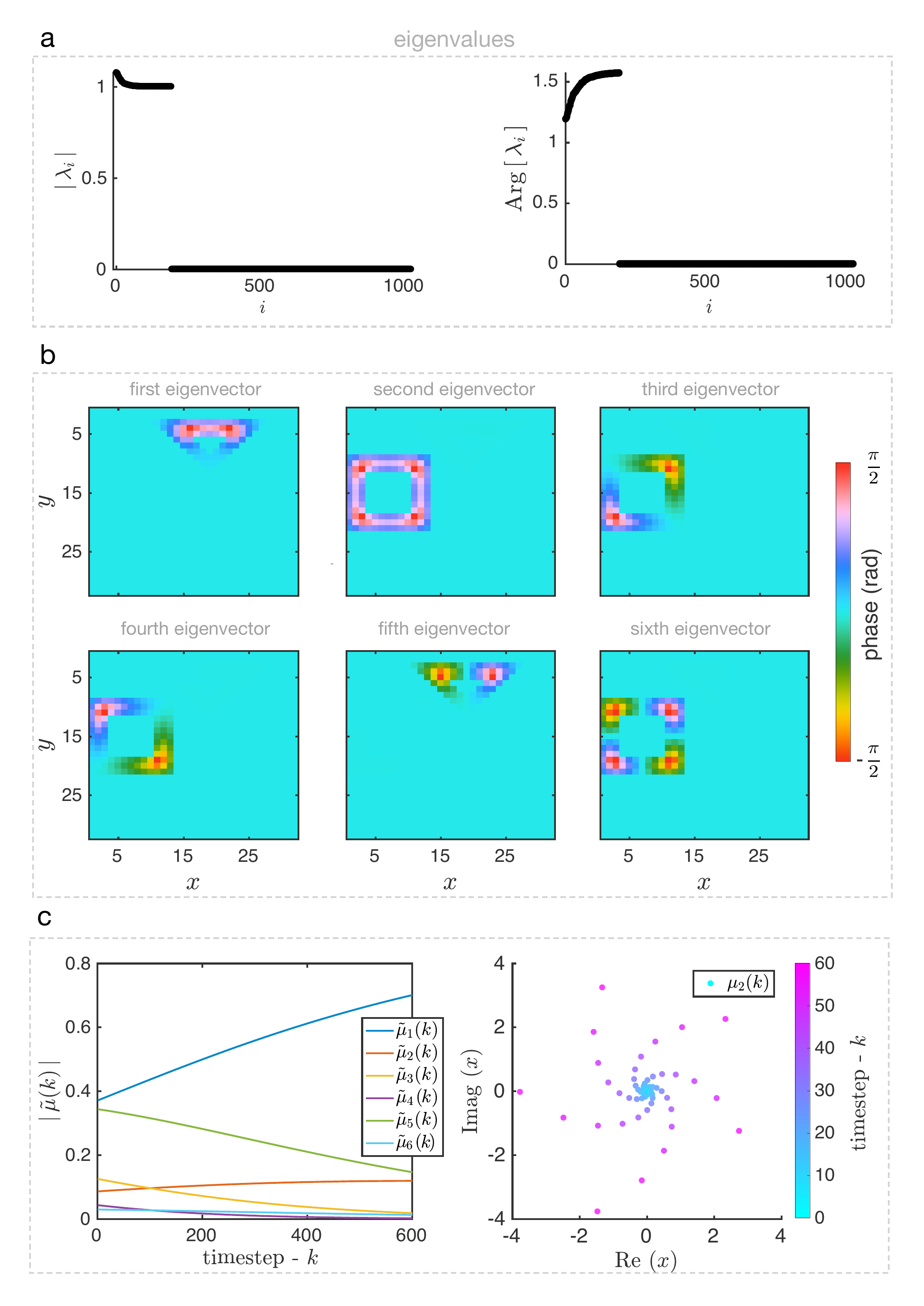}
    \caption{\textbf{Our approach offers a precise mathematical analysis for the segmentation task}. \textbf{(a)} We plot the amplitudes and phases of the eigenvalues of $\bm{B}_2$. A finite number of eigenvalues have noticeably higher amplitudes. \textbf{(b)} We then consider the phase configuration of the six eigenvectors of $\bm{B}_2$ associated with the six leading eigenvalues. Here, we plot the phases of these eigenvectors reshaped to a $N \times N$ grid. The phases of each element of the eigenvectors shape the foreground objects, and their linear combination generates the spatiotemporal patterns. \textbf{(c)} The contribution of each eigenvector is quantified by $\mu_i(k)$ (Eq.~\eqref{eq.contr}), which depend on the initial condition of the system. Further, the eigenvalues $\lambda_i$ rotate the phase values of the eigenvectors' elements as time progresses, and an example for the second mode can be observed.}
    \label{fig:fig6}
\end{figure}

Figure~\ref{fig:fig6} shows the eigenvectors associated with the leading eigenvalues of the second layer of the segmentation case in Figure 3a. We also show the normalized contribution:
\begin{equation}
\tilde \mu_{i} (k) =  \frac{| \mu_{i}(k) |}{\sum_i |\mu_{i}(k)|}
\label{eq.contr}
\end{equation}
of each eigenvector (Fig.\,\ref{fig:fig6}c, left panel), and depict the trajectory of contributions $\mu_{i}(t)$ (Fig.\,\ref{fig:fig6}c, right panel). Because we do not constrain the absolute values of the state vector, $\mu_{i}(k)$ increases or decreases asymptotically. Clustering the spatiotemporal patterns generated by the cv-RNN during the transient dynamics, however, is sufficient to achieve the object segmentation task. Finally, the spatiotemporal patterns produced by the linear complex dynamical system can be reproduced by reconstructing the dynamics using a linear combination of only a few eigenvalues and eigenvectors (see Supplementary Material, Sec. V and Supplementary Movie 6). Very differently from approaches that apply deep representations and intricate learning algorithms to accomplish object segmentation, this result demonstrates that the phase dynamics within a period of interest can be calculated very efficiently using low-rank approximations for the dynamics, which eases the computational burden of the task.

Neural network models are widely considered ``black boxes'', and current explainable AI methods~\cite{medicine2018opening, guidotti2018survey}, which aim to provide some rationale for the decisions and predictions made by a model, do not necessarily reveal the inner workings of neural networks. The mathematical formulation used in this work goes beyond offering heuristic insights into the model behavior, providing a precise mathematical equation and closed-form solution enabling complete interpretability of the mechanisms used for the computations. Using a single set of recurrent weights, object segmentation can be generalized across several different image inputs.

\subsection*{Discussion}

In this work, we have introduced a new recurrent neural network, with complex-valued dynamics, that can perform object segmentation in a range of images with a single set of weights, and that admits an exact solution. By focusing on transient dynamics and on connectivity regimes that lead to specific eigenvalue sets for the recurrent connectivity matrix $\bm{A}$, we find that the cv-RNN can perform segmentation using the simplest linear dynamics possible. This system, in turn, admits an exact mathematical solution, which we can leverage not only to explain exactly how this network performs the computation of image segmentation, but will also allow us to design dynamics to achieve new computations in future work. In this way, this linear cv-RNN represents an opportunity to drastically simplify some neural network computations in computer vision and beyond. Further, these results represent an advance in explainable artificial intelligence (XAI). The ability to specify the dynamics leading to the image segmentation computation in a precise mathematical expression surpasses current techniques for explaining how neural networks make decisions, and this mathematical approach may represent an important future direction for XAI research, specifically in introducing highly transparent and interpretable neural networks for computer vision and beyond. 

Previous work has studied oscillator networks to segment images through groups of nonlinear oscillators that synchronize when they are part of the same object \cite{ricci2021kuranet}. In addition to this example, other studies have also employed dynamical systems for image segmentation, such as the research conducted by \cite{belongie1998finding}, where the authors present an algorithm to find boundaries in natural images analogous to a spring-mass harmonic oscillator. In \cite{belongie1998finding}, Belongie and Malik developed a link between standard image segmentation algorithms such as normalized cuts to the dynamics of harmonic oscillators, by mapping the normalized cuts directly onto the eigenvectors of a harmonic oscillator system. This system corresponds to shaping the dynamics of a spring-mass oscillator network by changing connections using filtered versions of the input. Our results provide fundamental insight into how these networks of oscillators proposed by~\citet{belongie1998finding} and \citet{ricci2021kuranet} can learn to segment objects in images ranging from simple geometric constructions to naturalistic inputs. Further, our results also provide insight into how complex-valued autoencoders recently introduced by \citet{lowe2022complex} learn to synchronize phases with each different object. In the end, our results provide a critical simplification: a complex-valued linear dynamical system can perform the segmentation computation in its transient dynamics, without the need for training algorithms. The fact that one connectivity matrix allows the system to segment many images, without learning a new structure of network connections, reframes the problem of image segmentation into a single, very specific recurrent neural network architecture. We can then analyze how the network performs this computation through a direct mathematical analysis of the eigenspectrum of the resulting system matrix.

Recent interest in RNNs has centered on a deeper understanding of their underlying mechanisms and strategic design choices, particularly by incorporating complex-valued activations \cite{orvieto2023universality, orvieto2023resurrecting}. These efforts have shown that linear RNN layers can exhibit remarkable expressive power when coupled with multi-layer-perceptron blocks. In fact, they have outperformed their nonlinear counterparts in tasks of long sequence prediction~\cite{orvieto2023universality,orvieto2023resurrecting}. The results in this present work, however, demonstrate that linear cv-RNNs can perform highly sophisticated computations without additional processing layers, such as multi-layer perceptrons, as one may initially expect. Our findings thus not only align with these recent works but also streamline the network architecture by showing that fully linear recurrent layers are sufficient for object segmentation, which is a central task in computer vision. The way the cv-RNN solves this problem is not susceptible to problems that often arise when training recurrent neural networks, such as the vanishing gradient problem \cite{bengio1993problem, Goodfellow-et-al-2016}. Our results, further, allow a mechanistic interpretability for these networks that greatly improves the understanding of their inner workings and could potentially contribute to the development of novel computational algorithms.

Taken together, these results demonstrate the utility of recent insights into oscillator network dynamics \cite{budzinski2022geometry,budzinski2023analytical} and their potential interdisciplinary application to computer vision tasks. The cv-RNN introduced here performs object segmentation through relatively simple network dynamics that can also be solved exactly. Because most segmentation algorithms require highly sophisticated training regimes, this approach has the potential to drastically reduce the computational burden of some image processing tasks. Further, in cases where reduced precision for segmentation can be tolerated, this cv-RNN admits a simple low-rank approximation that can be easily truncated at any order. The flexibility of this approach demonstrates, in an additional manner, the utility of having a comprehensive mathematical description for neural networks. We hope that this first example of a neural network that can both perform a non-trivial computer vision task and be solved exactly opens doors for application across domains while also leading to innovative new algorithms in the field of image segmentation.

\section*{Code availability}

An open-source code repository for this work is available on GitHub: \href{http://mullerlab.github.io}{\textcolor{Cerulean}{http://mullerlab.github.io}}.

\begin{acknowledgments}
This work was supported by BrainsCAN at Western University through the Canada First Research Excellence Fund (CFREF), the NSF through a NeuroNex award (\#2015276), the Natural Sciences and Engineering Research Council of Canada (NSERC) grant R0370A01, Compute Ontario (computeontario.ca), Digital Research Alliance of Canada (alliancecan.ca), the Western Academy for Advanced Research, and NIH Grants U01-NS131914 and R01-EY028723. RCB gratefully acknowledges the Western Institute for Neuroscience Clinical Research Postdoctoral Fellowship. 
\end{acknowledgments}


%


\renewcommand\thefigure{M\arabic{figure}}
\renewcommand\theequation{M\arabic{equation}}
\setcounter{figure}{0}
\setcounter{equation}{0}

\end{document}